\documentclass[a4paper,twoside]{article}
\usepackage[utf8]{inputenc}
\usepackage[T1]{fontenc}
\usepackage{epsfig}
\usepackage{subfigure}
\usepackage{calc}
\usepackage{amssymb}
\usepackage{amstext}
\usepackage{amsmath}
\usepackage{amsthm}
\usepackage{svg}
\usepackage{multicol}
\usepackage{pslatex}
\usepackage{apalike}
\newcommand\Tstrut{\rule{0pt}{2.6ex}}       
\newcommand\Bstrut{\rule[-0.9ex]{0pt}{0pt}} 
\usepackage{SCITEPRESS}     

\begin{document}

\title{Measuring the Data Efficiency of Deep Learning Methods}

\author{\authorname{Hlynur Davíð Hlynsson, Alberto N. Escalante-B., Laurenz Wiskott}    
\affiliation{Ruhr University Bochum, Universitätsstraße 150, 44801 Bochum, Germany}
}

\keywords{Data Efficiency, Deep Learning, Neural Networks, Slow Feature Analysis, Transfer Learning}

\abstract{In this paper, we propose a new experimental protocol and use it to benchmark the data efficiency --- performance as a function of training set size --- of two deep learning algorithms, convolutional neural networks (CNNs) and hierarchical information-preserving graph-based slow feature analysis (HiGSFA), for tasks in classification and transfer learning scenarios. The algorithms are trained on different-sized subsets of the MNIST and Omniglot data sets. HiGSFA outperforms standard CNN networks when the models are trained on 50 and 200 samples per class for MNIST classification. In other cases, the CNNs perform better. The results suggest that there are cases where greedy, locally optimal bottom-up learning is equally or more powerful than global gradient-based learning.}

\onecolumn \maketitle \normalsize \vfill

\section{\uppercase{Introduction}}
\label{sec:introduction}

\noindent In recent years, we have seen convolutional neural networks (CNN) dominate benchmark after benchmark for computer vision since the 2012 ImageNet competition breakthrough \cite{krizhevsky2012imagenet}. These methods prosper with an abundance of labeled data, and an abundance of data is often required for acceptable results \cite{oquab2014learning}. In contrast, for most people, it is only necessary to see one picture of an Atlantic Puffin to be able to identify correctly such a bird as one. \\
\indent To be fair, we have a lot of prior experience. It is easy to make a mental note: ``a puffin is a small black and white bird with orange feet and a colorful beak'' because we have learned a useful representation of the salient aspects of the image. Instead of being bogged down by the details of every exact pixel value, as an untrained AI might, we can focus our attention on the most useful features of the image. \\
\indent For this reason, investigations on the efficacy of methods to learn a concept from few samples are often done through the lens of representation learning \cite{bengio2013representation}, for example via transfer learning \cite{pan2010survey} or low-shot learning \cite{wang2018low}. \\
\indent In this work we consider a method to measure data efficiency, the performance of an algorithm as a function of the number of data points available during training time, which is an important aspect of machine learning \cite{kamthe2017data},  \cite{al2015efficient}. We quantitatively examine the performance of CNNs and hierarchical information-preserving graph-based slow feature analysis (HiGSFA) \cite{escalante2016improved} networks for varying training set sizes and for varying task types. \\
 \indent HiGSFA has been chosen because it is the most recent supervised extensions of slow feature analysis (SFA) and has shown promise in visual processing with a notable distinction from CNNs: the computation layers are trained in a "greedy" layer-wise manner instead of via gradient descent \cite{escalante2016improved}. \\
 \indent The methods are applied to  visual tasks: a simple version of the MNIST classification task, where we vary the number of training points, and increasingly difficult tasks constructed from the Omniglot dataset. Our \textbf{contribution} in this work is a novel experimental protocol for evaluation of transfer learning applied to experimentally evaluate CNNs with the slowness-based HiGSFA.

\section{\uppercase{Related Work}}

\noindent Gathering data can be quite costly, so the question "how much is enough" has been considered in literature ranging from classical statistics \cite{krishnaiah1980handbook} over pattern recognition \cite{raudys1991small} to experimental design \cite{beleites2013sample}.
As data plays a central role in machine learning as well, the study of its effective use has garnered attention from all branches of the field. \\
\indent In a similar vein as our work, \cite{lawrence1998size} analyze the effect of generalization when the number of sample points are varied for supervised learning tasks. Equipped with the prior that supervised learning methods' performance obeys the inverse power law, \cite{figueroa2012predicting} trained a model to predict the classification accuracy of a model given a number of inputs.\\
\indent Transfer learning straddles the intersection between supervised learning and unsupervised learning, where the focus is uncovering representations that are both general and also useful for particular applications. The  Omniglot data set we consider was introduced in \cite{lake2015human} and has been popular  for developing transfer learning methods \cite{bertinetto2016learning}, \cite{edwards2016towards}, \cite{schwarz2018progress}.\\ 
\indent
With its sparse rewards and problems of credit assignment, reinforcement learning (RL) has a particular need for data efficiency, motivating such early works as prioritized sweeping \cite{moore1993prioritized}. More recently, \cite{riedmiller2005neural} designed the neural-fitted Q-learner for data efficiency. This method has been successfully combined with deep auto-encoder representations for visual RL \cite{lange2010deep}. Deep Q-Networks have made better still use of data for RL by combining experience replay, target networks, reward clipping and frame skipping \cite{mnih2013playing} \cite{mnih2015human}.\\
\indent
SFA was introduced in 2002 by Wiskott and Sejnowski as an unsupervised learning method of temporally invariant features \cite{wiskott2002slow}. These features can be learned hierarchically in a bottom-up manner, reminiscent of deep CNNs: slow features are learned on spatial patches of the input and then passed to another layer for slow feature learning.  The method is then called hierarchical slow feature analysis (HSFA) and has attracted attention in neuroscience for plausible modeling of grid, place, spatial-view, and head-direction cells  \cite{franzius2007slowness}. \\
\indent For labeled data, the method admits a supervised extension in the form of graph-based SFA (GSFA)  \cite{escalante2013solve}. Information is often lost in early layers of hierarchical SFA --- that could contribute to a globally slower signal --- prompting the development of HiGSFA  \cite{escalante2016improved}.\\
\indent Deep learning extensions of SFA is currently an active research area. The SFA problem is  solved with stochastic optimization in Power-SFA  \cite{schuler2018gradient}. A differentiable whitening layer is constructed, allowing for a non-linear expansion of the input to be learned with backpropagation. Another recent method, SPIN \cite{DBLP:journals/corr/abs-1806-02215} learns eigenfunctions of linear operators with deep learning methods and can be applied to the SFA problem as well.

\section{\uppercase{Methods}}

\noindent Below we describe the novel experimental setup as well as the methods being evaluated using the setup. For the remainder of the article we assume CNNs to be well-known and understood but we can recommend  \cite{cs231n2017convolutional} as a good pedagogical introduction to the method.

\subsection{HiGSFA}

HiGSFA belongs to a class of methods motivated by the slowness principle, which is based on the assumption that important aspects vary more slowly than unimportant ones \cite{sun2014dl}. This model takes as input data points such that data point $x_n$ is node $n$   in an undirected graph with weight $v(n)$. This can control the relative weight each data point has during the training but we set it as uniformly 1 is our experiments below.

The edge between nodes $n$ and $n'$ is $\gamma_{n, n'}$ and signifies a relationship between the data. This could be their spatial or temporal proximities or whether they belong to the same class.

For instance, during our classification tasks below,  we set:

\begin{equation}
    \gamma_{n, n'}= 
\begin{cases}
    1,& \text{if } n \text{ and } n' \text{ in same class}\\
    0,              & \text{otherwise}
\end{cases}
\end{equation}

Given a function space $\mathcal{F}$ with elements $g_j$, we learn slowly varying features $y_j(n) = g_j(x_n)$ of the data by solving the optimization problem \cite{escalante2013solve}:

\begin{equation}
\begin{aligned}
& \underset{g_j}{\text{minimize}}
& & \frac{1}{R} \gamma_{n, n'} \sum\limits_{n, n'} \left( y_j(n) - y_j(n') \right)^2 \\
& \text{subject to}
& & \frac{1}{Q} \sum\limits_{n} v_n y_j(n)\ \  \ \ \ \ \ \, = 0   \ \ \ \ \ \ \ \ \ \,  \\
&&& \frac{1}{Q} \sum\limits_{n} v_n \left(y_j(n) \right)^2\, \ \ \ = 1  \ \ \ \ \ \ \ \ \ \ \, \\
&&& \frac{1}{Q} \sum\limits_{n} v_n y_j(n) y_{j'}(n) = 0\text{, } j' < j \ \\
& \text{where}
&&Q = \sum\limits_{n}     v_n,\ R = \sum\limits_{n, n'} \gamma_{n, n'}
\end{aligned}
\end{equation}

The first constraint secures weighted zero mean, the second constraint secures weighted unit variance and the third one secures weighted decorrelation and order. \\
\indent To reduce computational complexity, we extract features of the data hierarchically. Similarly to CNNs, we extract features from $F \times F$ patches of the image data in the first layer, then extract features of $F' \times F'$ patches of the output features in the next layer and so on. The layers are trained by solving the optimization problem, one layer at a time, from the input layer to the output layer. The layer-wise parameters can be shared.

As we can experience information-loss while doing these layer-wise optimizations, an information-preserving mechanism is added. The cost function is minimized locally, so we can experience information-loss if dimensions are discarded that do not minimize the function on a local level --- but could conceivably be better for the overall problem.

\begin{figure}[!h]
  \fontsize{31}{05}\selectfont
  \centering
   {\resizebox*{0.4 \textwidth}{!}{\includegraphics
    {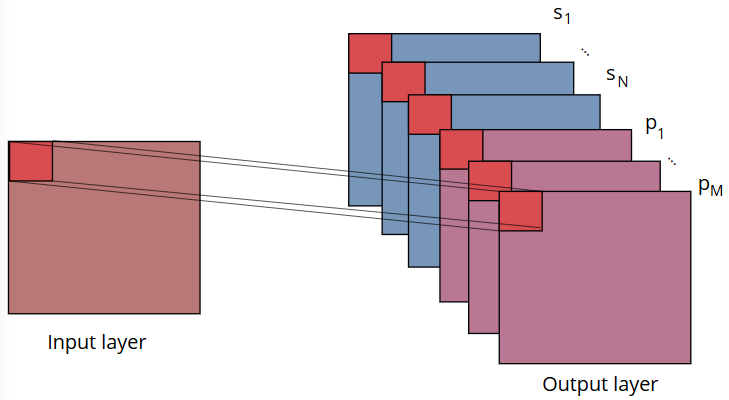}}}
  \caption{HiGSFA network layer. The feature generation is similar to that of the CNN. The layer outputs $N$ channels of slow features and $M$ channels of PCA features. The number of PCA channels features is either fixed beforehand or determined by replacing a number $M$ of the SFA features whose \textit{slowness} (cost function in eq. 2) exceeds a given threshold.}
  \label{example}
 \end{figure}

For each layer (figure \ref{example}), a threshold is placed on the features with respect to their slowness. If an output feature or features would be too fast, we replace them by the most variance-preserving PCA features. Each layer thus outputs a combination of slow features and PCA features.

 \subsection{General description of protocol}

The performance of two hypothesis $h_1$ and $h_2$, not necessarily from the same hypothesis set $\mathcal{H}$, is compared on a classification task. The learning curves of the two hypothesis are plotted as a function of the number of data points in the training set. This can be done simply by taking an increasing number of training points per class as we evaluate using MNIST, below. \\
\indent Alternatively, the number of training points per class are kept constant and the number of classes are varied. The relationship training and test set distributions is also altered, such that the task ranges from classical classification to transfer learning. We report a comparison of methods below using this scheme on the Omniglot data set.

\subsection{Evaluation on MNIST}
First, we compare classification accuracies on MNIST  \cite{lecun1998gradient} as a function of the number of samples per class used during training. The images have a dimension of $28 \times 28$ pixels.  For 100 iterations, we choose random samples from each class and use a thousand unused samples from each class for validation. Finally, the models are tested on the classic 10 thousand test images.

\subsubsection{Architectures}

We constructed a two-layer HiGSFA network with circa 13k parameters (the number is stochastic and changes from training set to training set), extracting 400 features from the data. The first layer has a filter size of $5\times5$ and a stride of 2, extracting 25 features for each spatial patch. The second layer has a filter size of $4\times4$ and a stride of 2, extracting 16 features for each spatial patch. \\
\indent The output of the first layer is concatenated with a copy of itself, where each element $x$ is replaced with $|x|^{0.8}$, doubling the number of channels and giving us nonlinearity.  If the value of the objective function is larger than a threshold of 1.99, we select PCA features. This upper bound is motivated by the fact that non-predictive, white noise features take a value of 2 in the objective function \cite{creutzig2008predictive}. The parameters within each layer are shared. A single-layer softmax neural network was trained on the features of the second layer to handle classification, which has 4010 parameters.\\
\indent Two standard CNNs were constructed as well, one with the constraint to have a similar number of parameters as the HiGSFA network, and another with an amount closer to what is seen in practice on similar datasets. That is to say, the smaller CNN corresponds to the HiGSFA network. \\
\indent We call the smaller network CNN-1 which has 10,032  trainable parameters, excluding the number in the final layer for classification. The tasks have varying numbers of classes to be predicted, causing the classification layer to have varying numbers of parameters. CNN-1 has three convolutional layers, each one followed by ReLU and max pooling, the first two with 8 channels and the last one with 16. They are followed by a fully connected classification layer, using a softmax activation function. The first convolutional layer has a filter size of $7\times7$, and the other two have a filter size of $5\times5$. The convolutional layers have a stride of 1 and the max pooling layers have a stride of 2.\\
\indent We call the larger network CNN-2, with 116,214 parameters (not counting the classification layer). It is the same as CNN-1 except the convolutional layers have twice the number of channels, and a dense layer with 150 units is added before the classification layer. \\
\indent Note that the parameter configurations of both HiGSFA and CNNs have not been optimized for the best performance on the tasks below. They were designed to be lightweight according to general best practices \cite{hadji2018we} \cite{escalante2016improved}. This allows for more trials and tighter confidence bounds while achieving fair performance on the tasks.
\subsection{Evaluation on Omniglot}
Omniglot is a handwritten character dataset consisting of 50 alphabets with 14 to 55 characters each, each character having 20 samples \cite{lake2015human}.  The alphabets vary from real alphabets, such as Greek, to fictional ones, such as Alienese (from the TV show “Futurama”). Each sample was drawn by a different person for this dataset. It is typically split into 30 training alphabets, and 20 testing alphabets. Note that the training-testing split separates the alphabets; all samples originating from all characters from a given alphabet appear in either the training set or the test set but not both. This makes it a transfer learning task as the training and test data set samples drawn from separate distributions.  

In the original work using the dataset, the methods were first trained on the 30 background alphabets, and then a 20 way one shot classification task was performed. Two samples are taken from each of 20 characters from random evaluation alphabets. One sample is placed in what we’ll call a probe set, and the other in a target set. The methods then try to find the corresponding sample in the target set that is the same character as any given sample in the probe set.

\begin{figure}[!h]
  \centering
   {\resizebox*{0.4 \textwidth}{!}{\includegraphics
    {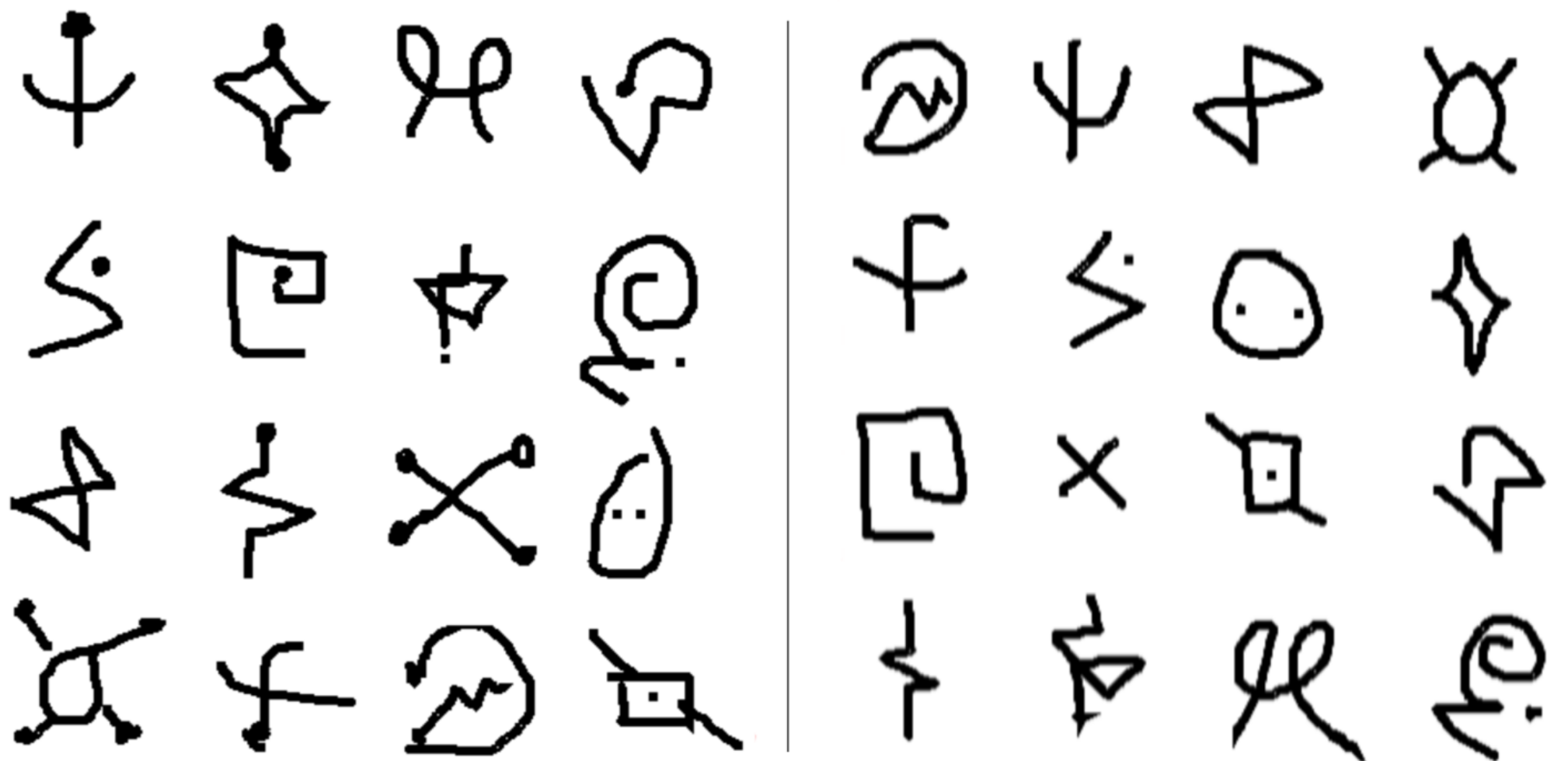}}}
  \caption{16 way one shot classification. Symbols on the left are presented to the algorithm, one at a time, and the task is to find the same character from the symbols on the right.}
  \label{fig-16way}
 \end{figure}

In the vein of the original Omniglot task, we compare several models in three challenges. In all challenges, we do 16 way one shot classification using 1-nearest-neighbor (1-NN) under the Euclidean distance. The challenges differ in how the test set is related to the training set:
\subsubsection{Challenge 0} From 16 random characters used for training, we take two samples that the models were trained on. These samples are placed in two sets, the probe and target sets, such that each set contains one sample of each character. The model under consideration extracts features from each image. We then iterate through each feature vector from images in the probe set and find the closest feature vector from the target set. If those two vectors belong to images of the same class, then we count it as a success.

\subsubsection{Challenge 1} Same as above, but we take characters used during training for the probe set and perform classification on samples that were not used during the training.

\subsubsection{Challenge 2} Same again, but now we do the classification on characters that do not belong to alphabets used during training.

\subsubsection{Omniglot architectures}

 All model architectures are the same for MNIST and Omniglot, but the Omniglot images are resized to $35\times35$, having the effect that HiGSFA outputs 784 features. The number of model parameters does not change as the weights are shared for the image patches. The HiGSFA features used for classification are simply the 784 output features and we do not train a neural network classifier on them.

The total number of parameters in the CNNs depend on the number of training classes, due to the classification layer. We fix the number of alphabets to 8 and vary the number of characters per class to be 4, 6, 8, 10, 12 and vice versa. The number of parameters for CNN-1 range between 18k and 35k, and for CNN-2 range between 121k and 130. If we do not count the parameters from the final classification layer, then the number of parameters for CNN-1 for these tasks is always 10,032  and the number for CNN-2 is 116,214.

After training the CNNs, we perform feature extraction by intercepting the output of the second-to-last layer. Here the assumption is that CNNs learn a representation for the classification layer \cite{DBLP:journals/corr/RazavianASC14}. We are then interested in comparing the strength of HiGSFA and CNN representations when used by a 1-NN classifier.

\begin{table*}[]
\centering
\begin{tabular}{  r | l l | l l | l l}

\hline
  Samples\,
 & HiGSFA & & CNN-1 && CNN-2 &\Tstrut  \\
         & Acc. & Std.    &  Acc.  & Std.  & Acc.  & Std.  \Bstrut     \\
\hline
    5\,     & 35.683    & $\pm$ 0.430  & $\textbf{72.361}$   &$\pm$ 0.365 & 72.320 &$\pm$ 0.094\,  \,\,\ \Tstrut\\
  10\,   & 75.736    &    $\pm$ 0.222  & $\textbf{80.392}$   &$\pm$ 0.241 & 79.551 &$\pm$ 0.175     \\
  50\,   & $\textbf{92.970}$    &$\pm$ 0.050  & 90.320   &$\pm$ 0.101   & 91.465 & $\pm$ 0.070    \\
  200\,  & $\textbf{96.246}$    &$\pm$ 0.027   & 94.672   &$\pm$ 0.062  & 95.648 & $\pm$ 0.051    \\
  500\,   & 97.188   &$\pm$ 0.013    & $96.579$   &$\pm$ 0.046  & \textbf{97.308} &$\pm$ 0.054  \\
  2000\,   & 97.887   &$\pm$ 0.009 & 98.247   &$\pm$ 0.020  & $\textbf{98.571}$ & $\pm$ 0.023   \\
  6000\,   & 98.134   & $\pm$ 0.008   & 98.687   &$\pm$ 0.014 & $\textbf{98.949}$ &$\pm$ 0.015  \\
\end{tabular}
\caption{\textbf{MNIST Accuracies.} The percentage of correctly classified samples on the test set along with the standard error of the mean (SEM). }
\label{tab:totalresults}
\end{table*}

\subsubsection{Training}

The models were trained on varying amounts of samples per character. The HiGSFA network was trained to solve the optimization problem on each image patch, one layer at a time.  All neural networks were trained in Keras  \cite{chollet2015keras} using ADAM  \cite{kingma2014adam}, with default parameters, to minimize cross-entropy.\\ \indent After each epoch, the error was calculated on the validation set. Early stopping was performed after the validation error had increased four times in total during the training. The training for Omniglot is the same, except instead of early stopping, the CNNs were trained for 20 epochs in all cases.

\begin{figure*}[!h]%
    \centering{{\includegraphics[width=5.7cm]{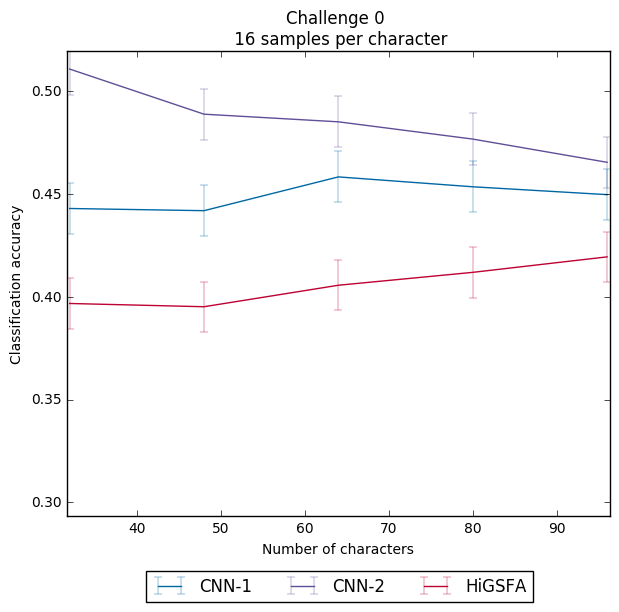} }}%
    \qquad{{\includegraphics[width=5.7cm]{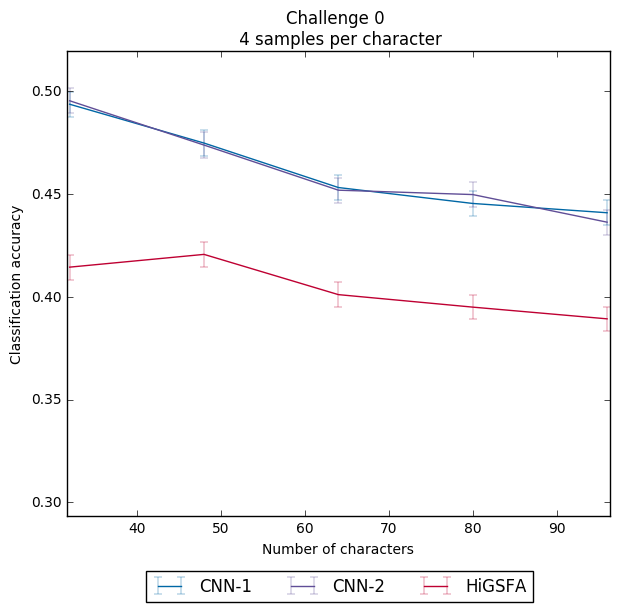} }}%
    \qquad{{\includegraphics[width=5.7cm]{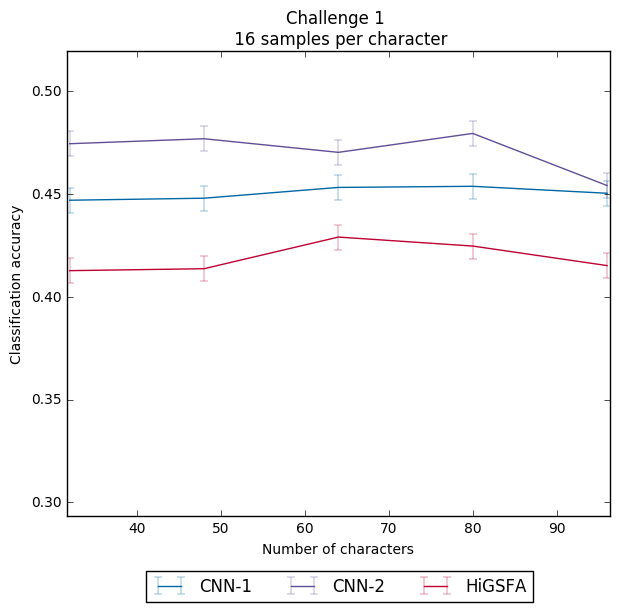} }}%
    \qquad{{\includegraphics[width=5.7cm]{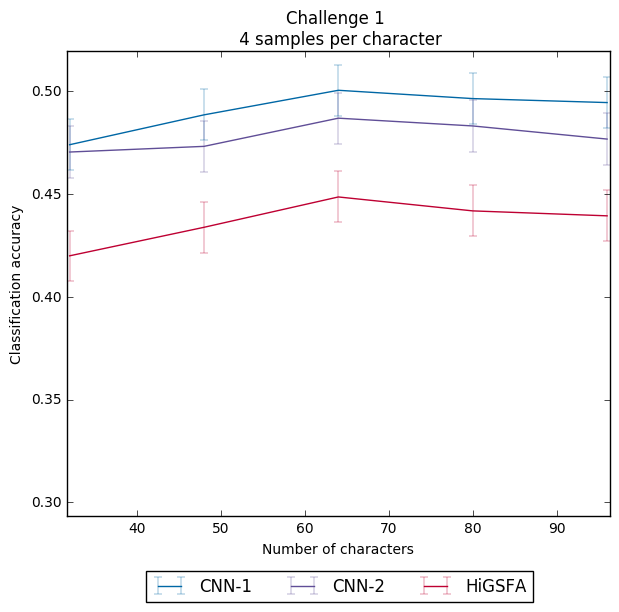} }}%
    \qquad{{\includegraphics[width=5.7cm]{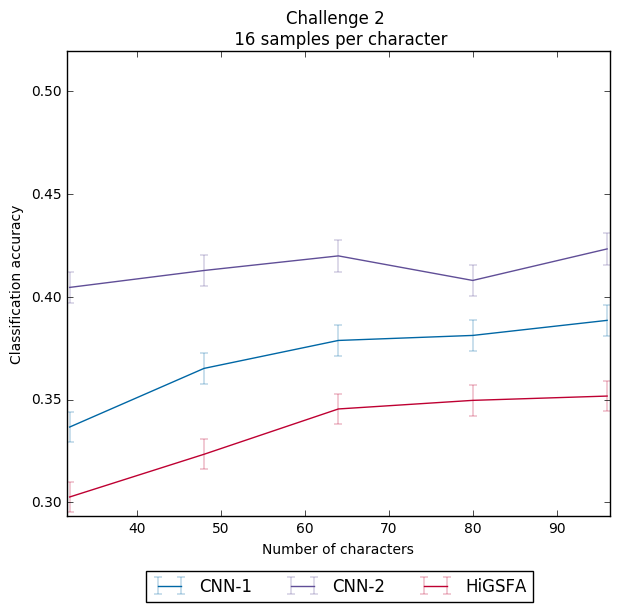} }}%
    \qquad
    {{\includegraphics[width=5.7cm]{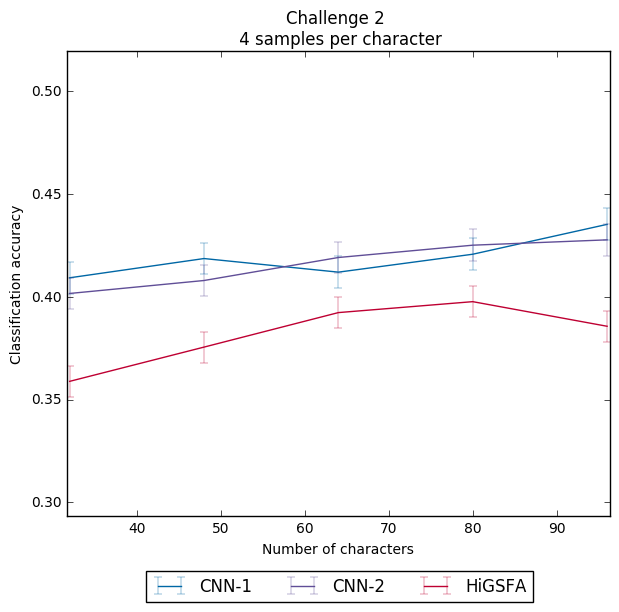} }}%
    \caption{\textbf{Classification Accuracies}. There are either 8 alphabets and we vary the characters per alphabet, or vice versa. The error bars indicate the standard error of the mean. These plots are best viewed in color.}%
    \label{fig:plots}%
\end{figure*}

\section{\uppercase{Results}}

\subsection{MNIST Results}
We trained the models using 5, 10, 50, 200, 2000 or 4000 samples per digit. In table \ref{tab:totalresults}, we see the statistics from 100 runs, where the models were trained from random initializations, evaluated and tested.  The convolutional networks have the highest accuracies when there are 2000 or more samples per class and when there are only 5 or 10 samples per class. \\ \indent However, HiGSFA has a higher accuracy than the CNN with a similar number of parameters for 500 samples per class. Furthermore, HiGSFA has higher accuracies than both CNNs for 200 and 50 samples per class. The CNN with a larger number of parameters always has higher prediction accuracies than the one with a lower number of parameters.

\subsection{Omniglot Results}

The 1-NN classifier uses the second-to-last CNN outputs or HiGSFA features. We fix either the number of alphabets, or characters-per-alphabet, to be 8 and vary the other number from 4 to 12 in increments of 2. The number of samples per character is either 4 or 16. The largest total number ($\text{alphabets} \times \text{characters per alphabet} \times \text{samples per character}$) of samples used for training is 1536 and the lowest is 128. \\
\indent In figure \ref{fig:plots}, we see the average of all the runs over the different samples per characters and number of classes. In all of the challenges, the CNNs have higher accuracies than HiGSFA. On average, CNN-2 has higher accuracies in challenges 0 and 2. Neither CNN achieves significantly better accuracy than the other in challenge 1.

\section{\uppercase{Discussion and conclusion}}

\noindent The work of this paper is intended to facilitate understanding of algorithms from the point of view of having particularly low numbers of samples. We present simple-to-implement challenges that allow for evaluation of data efficiency in the context of representation learning. \\
\indent For the models experimented on, we see that the CNNs usually perform better, but HiGSFA outperforms the CNNs on 50 and 200 sample training sets from the MNIST data. One can speculate that the default CNN architectures ensure generalization through max-pooling whereas SFA mostly learns to generalize from a moderately sized data set. \\
\indent Another explanation for the different ranges of comparative performance optima is the choice of delta-threshold of HiGSFA. The method overestimates the slowness of the slowest features when it has too few samples. This has the effect that fewer PCA features are selected for a lower number of samples. On the other hand, with more than 200 samples, there could be too many PCA features chosen. Setting the number of slow features to be a constant for all sample sizes could be better for robustness than fixing the delta threshold. \\
\indent Notice the trend in challenge 0: the accuracy goes down as the number of samples increases. This is due to the samples used for the probe and target sets being drawn from the training set and we are training and testing on larger sets as we move from left to right. \\
\indent Overall, for the Omniglot challenges, the accuracies of the CNNs lie comfortably above the HiGSFA accuracies, but it’s not always discernible whether the larger or the smaller CNN performs better. An explanation for this could be that the tasks are not difficult enough for more parameters to be necessary.  The local optimality of GSFA could result in an insufficiently robust or transferable representation if there are many classes and few samples per class. \\ 
\indent These challenges are more complicated set of classification tasks than the MNIST ones, with a larger number of classes overall. This give CNNs an opportunity to take advantage of having been trained directly for classification when they are presented a similar task. Although HiGSFA takes advantage of class labels, it suffers in comparison for not taking into account the downstream task during training.\\
\indent For future work, a complete extension of the experiments here could include an analysis on the effect that different type of data would have on the performance. This would yield further insight than varying the number of rather homogeneous data used for training. Additionally, the performance of a wider array of popular methods can be compared. \\
\indent More types of benchmarks for comparing different models over varying training set sizes would be helpful for this kind of research. Knowledge gained from them would as well allow practitioners to choose the right model for the scale and type of the problem they wish to solve.  These experiments give rise to the question: how can these methods with their different strengths and weaknesses profit from each other?

\vfill
\bibliographystyle{apalike}
{\small
\bibliography{example}}

\vfill
\end{document}